\documentclass{article}
\usepackage{spconf,amsmath,graphicx,balance}

\title{TARGET-QUALITY IMAGE COMPRESSION WITH RECURRENT, CONVOLUTIONAL NEURAL NETWORKS}
\name{M. Covell, N. Johnston, D. Minnen, S.J. Hwang, J. Shor, S. Singh, D. Vincent, G. Toderici}
\address{Google Research, 1600 Amphitheatre Parkway, Mountain View CA 94043}
\begin{document}
%
\maketitle
\begin{abstract}

We introduce a stop-code tolerant ({\em SCT}) approach
to training recurrent convolutional neural networks for lossy image
compression. Our methods introduce a multi-pass training method to
combine the training goals of high-quality reconstructions in areas
around stop-code masking as well as in highly-detailed
areas. These methods lead to lower true bit\-rates for a given
recursion count, both pre- and post-entropy coding, even using
unstructured LZ77 code compression.  The pre-LZ77
gains are achieved by trimming stop codes.  The
post-LZ77 gains are due to the highly unequal distributions of 0/1
codes from the SCT architectures.  With these code compressions, the
SCT architecture maintains or exceeds the image quality at all
compression rates compared to JPEG and to RNN auto-encoders across the Kodak
dataset.  In addition, the SCT coding results in lower variance in
image quality across the extent of the image, a characteristic that
has been shown to be important in human ratings of image quality.

\end{abstract}
\section{INTRODUCTION}
\label{sec:intro}

Earlier work has shown the power of convolutional neural networks in
compressing images, both under a single-bitrate
target~\cite{BalleLS16a} and under multiple-bitrate
targets~\cite{Toderici2016,gregor2016conceptual}.  Both approaches are
better than JPEG compression,
as long as entropy coding is used on their output
symbols~\cite{BalleLS16a,Toderici2016}.
To date, both types of neural networks have suffered from the
constraint that their output codes have a fixed symbol dimension (``depth'')
over the full extent of the image.  In~\cite{BalleLS16a}, the number
of symbols
is completely defined by the depth of the bottleneck layer.
Similarly,
\cite{Toderici2016} allows for different
quality levels, by changing the number of iterations, but the number
of iterations is fixed across the full image, giving an equal number
of symbols to each area of the image.
We seek
to send fewer symbols in simpler sections of the image, allowing us
to send additional symbols on more difficult-to-compress
sections.  This allows neural-network based systems to see
 gains similar to whose from
run-length encoding in JPEG DCT
coefficient transmission.

\begin{figure}
  \centering
  \begin{minipage}{0.49\columnwidth}
  \centering       
  \includegraphics[width=\linewidth]{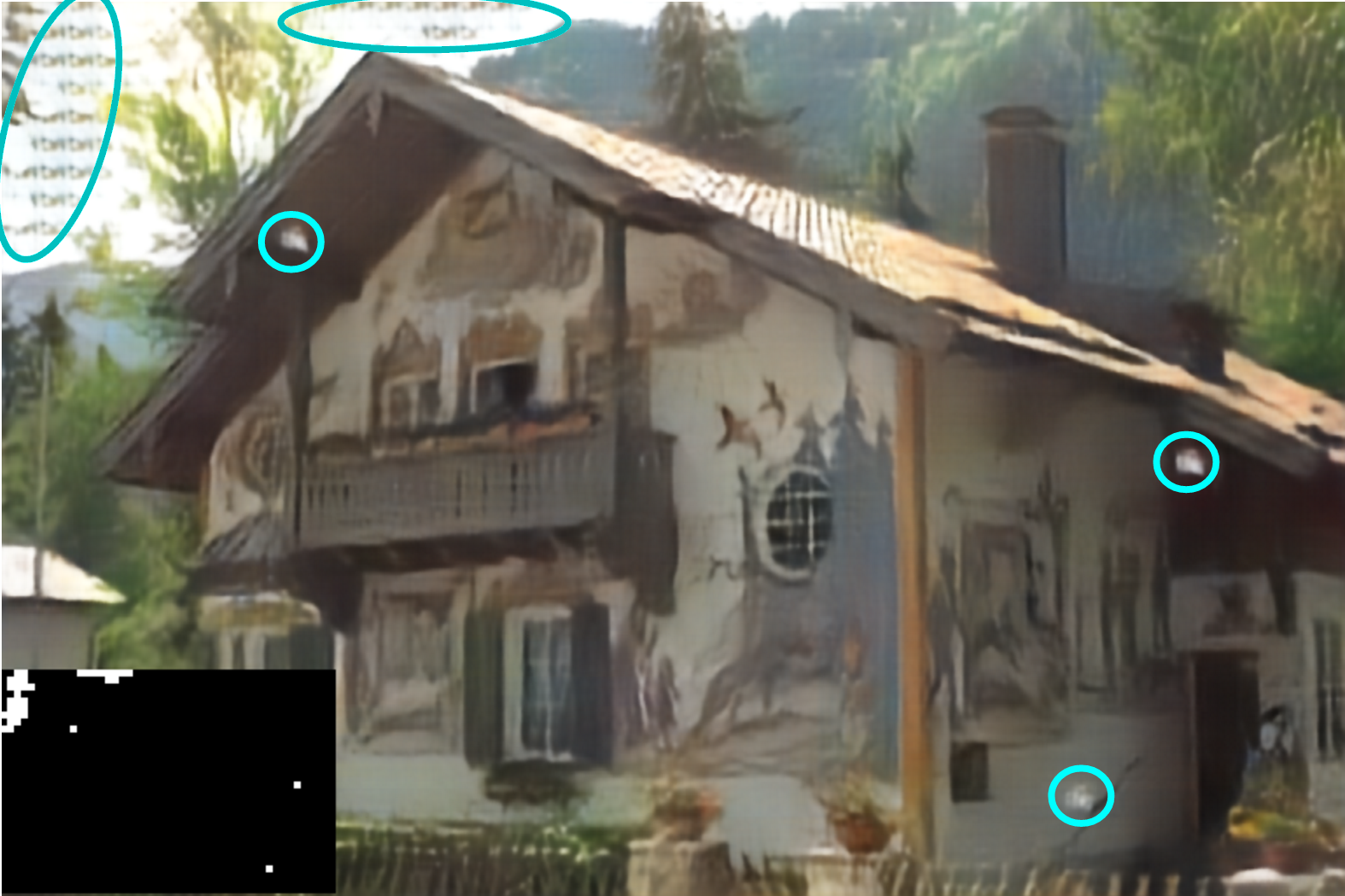}
  \linebreak \small
  (a) 0.121 bpp non-SCT
  \end{minipage}
  \hfill
  \begin{minipage}{0.49\columnwidth}
\centering
\includegraphics[width=\linewidth]{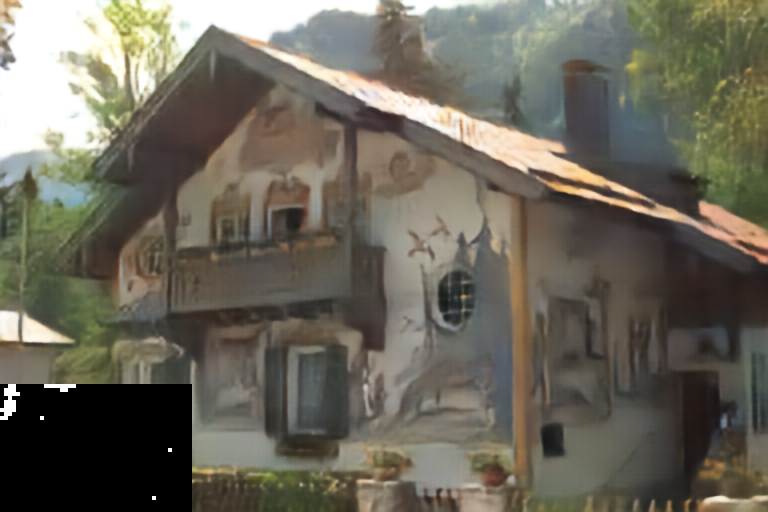}
  \linebreak \small
  (b) 0.115 bpp SCT
  \end{minipage}

\caption{\small Examples of recursive compression at 0.125 bpp (nominal),
  with the stop-code mask in the lower-left
  corner.  (a) Output from LSTM, additive, adaptive-gain network
  trained as described in~\cite{Toderici2016}; after gzip coding this
  is 0.121 bpp. Code-trimming-induced artifacts are circled in cyan.  (b) Same
  architecture but trained as described in
  Section~\ref{sec:method}; after gzip coding this
  is 0.115 bpp.  Using this modified training, we avoid artifacts from
  code trimming.}
  \label{fig:results-est-example-iter0}
\end{figure}

The difficulty in stop codes and code trimming in recursive neural
networks ({\em RNNs}) is 
their convolutional nature results in wide spatial dependence on
symbols that are omitted (due to earlier stop codes).
Simple training approaches for stop-code tolerant ({\em SCT}) RNNs
tend to produce blocking artifacts  and blurry reconstructions.
The blocking artifacts in the areas around the stop-code-induced gaps
(Figure~\ref{fig:results-est-example-iter0}) are due to the
RNN's convolutions
relying on neighbor codes too heavily and then failing when
those codes are omitted.
Blurry reconstructions
are due to the learning process attempting to
accommodate code trimming even in complicated image patches.

This paper introduces a two-pass approach to training RNN's for
high-quality image reconstruction with stop codes and code trimming:
On each iteration, the first pass trains the decoder
network with examples that include stop-code masking and the second
trains for accurate reconstructions.  By introducing the first pass,
the full training process (both early and late) includes the SCT
requirement, allowing the network to converge to symbol assignments
that both satisfy the stop code structure and provide representational power.

\begin{figure}
  \centering
  \begin{minipage}{\linewidth}
  \centering   \small
  \includegraphics[width=\linewidth]{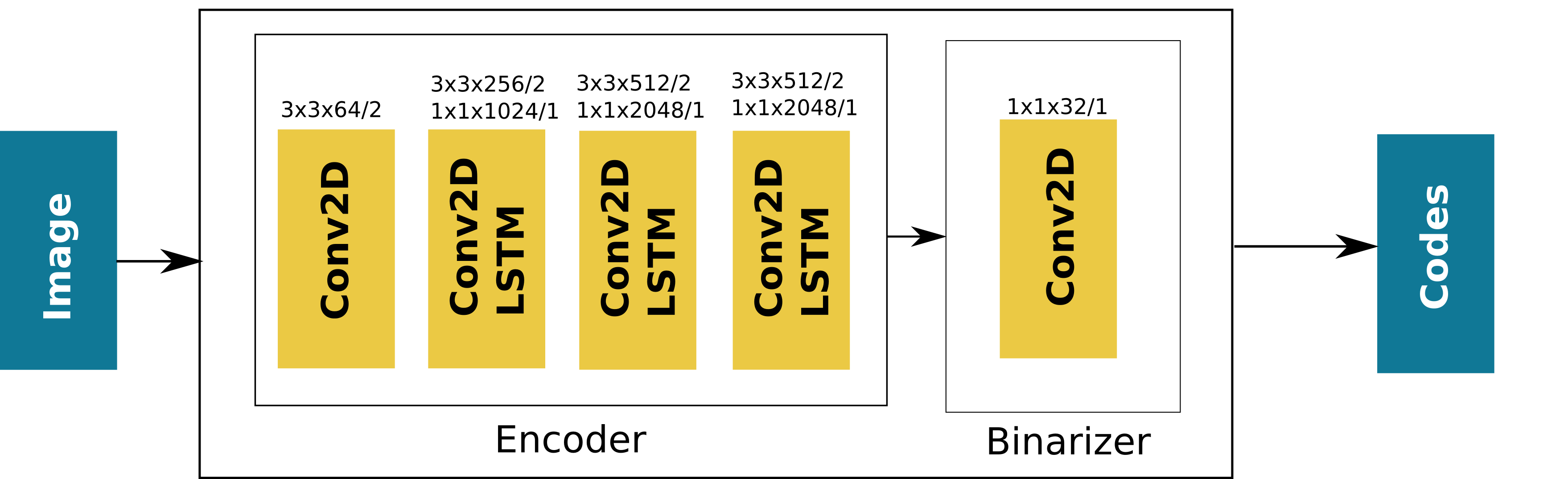}
  \linebreak \small
  (a) Encoder and binarizer used by both non-SCT and SCT networks.
  \end{minipage}
  \begin{minipage}{\linewidth}
  \centering   \small
  \includegraphics[width=\linewidth]{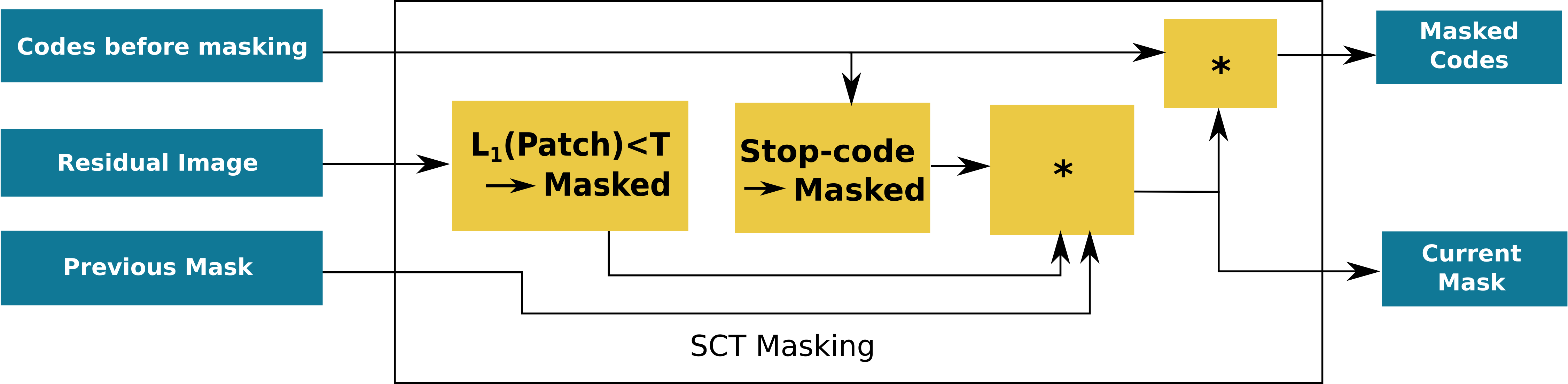}
  \linebreak \small
(b) SCT-network masking logic, to enforce stop-code behavior.
  \end{minipage}
  \begin{minipage}{\linewidth}
  \centering   \small
  \includegraphics[width=\linewidth]{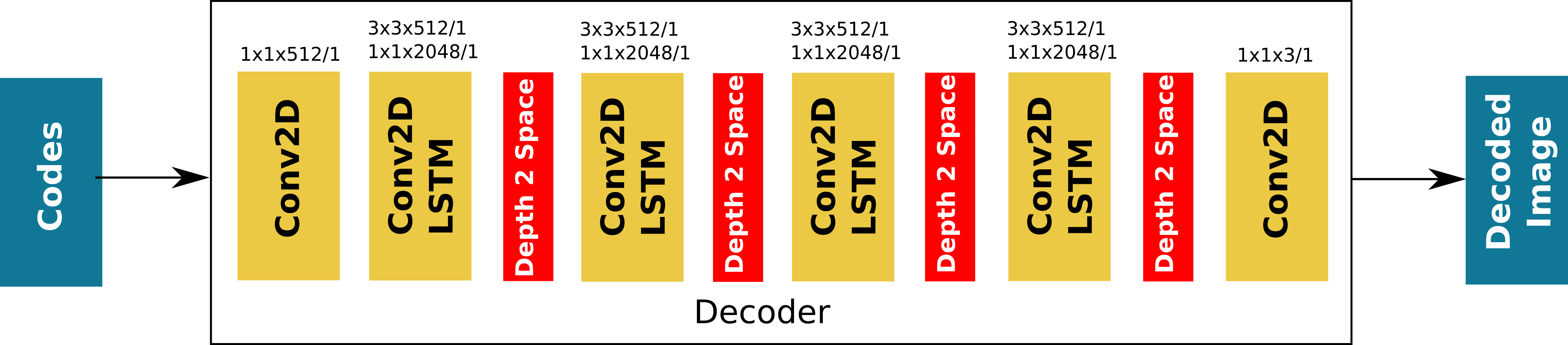}
  \linebreak  \small
(c) Decoder used by both non-SCT and SCT networks.
  \end{minipage}
  \caption{\small Block diagram for both non-SCT (by omitting {\em b}) and
  SCT (by including {\em b}). Parameters for the
  convolution layers are listed as HxWxD/S, for kernels that are HxW with
  D channels and a stride of S. Convolutional LSTMs list input convolutions
  above hidden-state convolutions.  The masking logic ensures that
  once a stop code is emitted (whether accidentally or deliberately),
  all subsequent iterations feed a stop code at that tile to the decoder.}
  \label{fig:block diagram}
\end{figure}

\section{PREVIOUS WORK}
\label{sec:previous}

Our research is based on the confluence of two long-running areas of
research: image compression~\cite{JPEG,WebP,BPG} and
neural-network-based `auto-encoders'
\cite{Hinton2006,Krizhevsky2011,vincent2010}.
Recurrent autoencoders have been
successfully
applied to the problem of variable-rate image compression 
\cite{Toderici2015, Toderici2016} while
feedforward neural networks have been equally successful at
fixed-rate encoding \cite{BalleLS16a}.  This previous
work in neural-network-based compression
implicitly allocates the same
number of symbols to each local `tile' of the image:
there was no equivalent to
code trimming as seen in JPEG~\cite{JPEG} via run-length encoding.

In this paper, we focus on training fully convolutional,
recurrent neural networks to handle stop codes.  A parallel
effort in this area~\cite{Minnen2017} looks at
solving the same data-adaptive coding problem by
moving to a block-based
approach, like JPEG and WebP. In contrast, this paper's work remains in the
fully-convolutional framework 
and 
establishes a predictible stop code, both to signal to
the decoder that a tile is entering code trimming and to allow the
decoder to 
fill in all subsequent unsent (stop) codes for that tile.

\section{METHOD}
\label{sec:method}

For our SCT approach, we start by using the
architecture of one of the best variable-rate-encoding architectures
from~\cite{Toderici2016}: {\em LSTM (Residual Scaling)}. In the
encoder network (Figure~\ref{fig:block diagram}-a), {\em LSTM
(Residual Scaling)} interleaves LSTM units
with four layers of spatial downsampling, giving the binarizer a
$\frac{H}{16} \times \frac{W}{16}$ representation of the $H \times W$
image.  In the decoder network (Figure~\ref{fig:block diagram}-c), a
full $H \times W$ reconstruction is
created from the $\frac{H}{16} \times \frac{W}{16}$ codes
using `depth-to-space' shuffling~\cite{tensorflow}.
In this paper, we
refer to this baseline as the ``non-SCT'' network.\footnote{The
non-SCT model used in this paper
trained for significantly longer than both
{\em LSTM (Residual Scaling)}~\cite{Toderici2016} and
our SCT model.
With equal
computational resources,
the SCT improvements are likely to increase from what is reported here.}
Figure~\ref{fig:block diagram} shows the architectures of our
non-SCT model (by excluding Figure~\ref{fig:block diagram}-b) and
our SCT model (by including Figure~\ref{fig:block diagram}-b).
As mentioned above, spatial downsampling
in the encoder results in the
binarizer outputting one stack of {\em bi-level} (-1/1) codes for each
$16 \times 16$
spatial `tile' from the input image but the
image-tile--to--code-stack mapping is not the same as the
blocks used in JPEG or BPG~\cite{BPG}: the convolutional support
of the encoder network results in spatial dependences far beyond
this downsampling-induced $16 \times 16$ association.

\begin{figure}
  \centering
  \begin{minipage}{0.45\columnwidth}
  \centering
  \includegraphics[width=\linewidth]{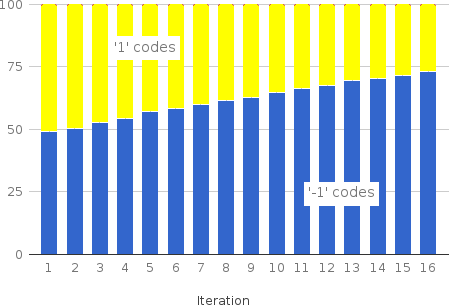}
  \linebreak \small
  (a) non-SCT -1/1 frequencies
  \end{minipage}
  \hfill
  \begin{minipage}{0.45\columnwidth}
  \centering
  \includegraphics[width=\linewidth]{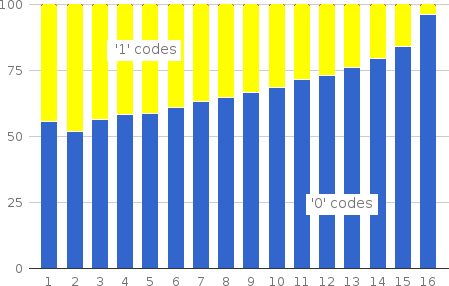}
  \linebreak \small
  (b) SCT 0/1 frequencies
  \end{minipage}
  \caption{\small The SCT architecture produces more biased code
  distributions than does the non-SCT, making LZ77 more effective,
  especially on later iterations.}
  \label{fig:bit-distributions}
\end{figure}

\begin{figure*}
  \centering
  \begin{minipage}{0.52\textwidth}
  \centering
  \includegraphics[width=\linewidth]{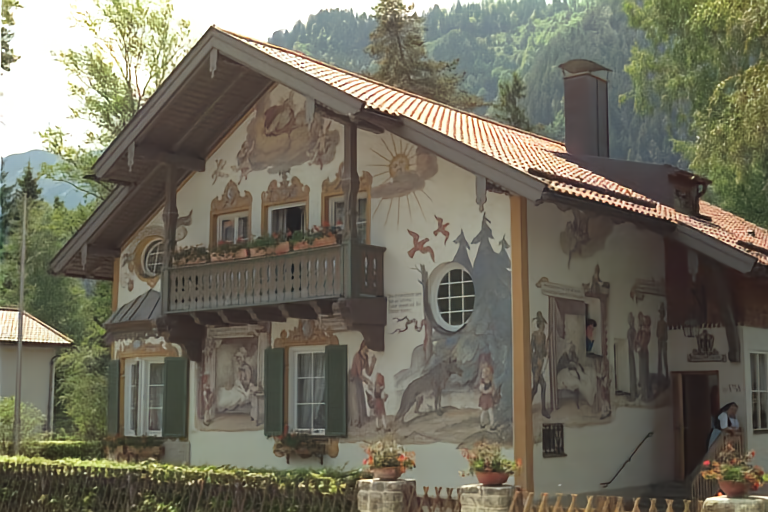}
  \linebreak \small
  (a) 0.791 bpp SCT reconstruction
  \end{minipage}
  \hfill
  \begin{minipage}{0.47\textwidth}
  \centering
  \begin{minipage}{0.49\linewidth}
  \centering
  \includegraphics[width=\linewidth]{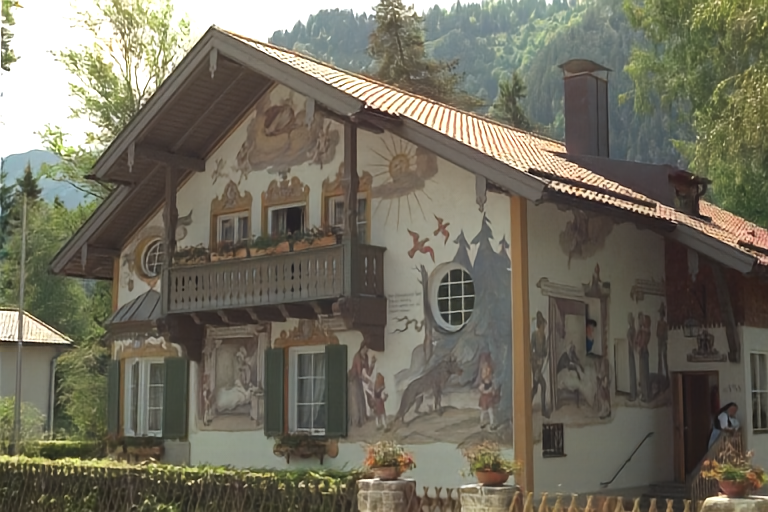}
  \linebreak \small
  (b) 0.745 bpp non-SCT
  \end{minipage}
  \hfill
  \begin{minipage}{0.49\linewidth}
  \centering
  \includegraphics[width=\linewidth]{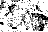}
  \linebreak \small
  (c) quality comparison
  \end{minipage}

\caption{\small Image reconstructions near 0.75 bpp. (a) Reconstruction
  from SCT architecture at 0.791 bpp ($7^{th}$ iteration which gives
  the closes true bpp to target). (b) Reconstruction from non-SCT
  architecture at 0.745 bpp ($6^{th}$ iteration). (c) Quality
  differences between SCT and non-SCT reconstructions.  SCT gives
  lower $L_1$ error on 65.0\% of the image (white); non-SCT gives
  lower error on 11.5\% of the image (black); and they are effectively
  equal ($\pm \frac{1}{2}$ out of 256) on 23.5\% of the image (gray).
}
  \label{fig:results-75bpp}
  \end{minipage}
\end{figure*}

In~\cite{Toderici2016}, the actual bit rates were measured by running the
binarizer codes through a custom-trained entropy coder.  For the sake
of time and clarity,
we instead use Lempel–Ziv coding
(LZ77)~\cite{gzip} on the flat file that contains the codes.
Note
that LZ77 compression is less effective than what is
possible with a custom entropy coder, since LZ77 does not exploit
the underlying spatial and depth structure of the codes.

We make minimal architecture modifications to go from the non-SCT
network to our SCT network.  We modify the binarizer ~\cite{Toderici2016} to use
0/1, instead of -1/1, to avoid boundary artifacts at the edges of images.
We also add a masking process after the binarizer layer, but before
``code transmission'' (Figure~\ref{fig:block diagram}-b), to enforce
our stop-code behaviors.  The mask is
computed at the end of each iteration, setting to all `0' (masked) those
tiles: (a)
where the reconstruction quality for the tile exceeds our target
quality level; (b) where the output of the encoder network for this iteration
is (accidently) the all-zero stop code; or
(c) where the codes were masked off on an earlier
iteration.
This masking logic allows us to treat an all-zero code
as a stop code and to avoid sending any of that tile's subsequent
iterations: the code itself communicates the stop condition.
Since the decoder is able to compute the mask for each iteration
without any additional information, we explicitly provide the mask
as input into the decoder.

The SCT behavior of our trained networks derives from the training
loss that we use.  Our training loss includes the
`reconstruction loss' used in~\cite{Toderici2016}.  If this is the
only term that is used (as it was in~\cite{Toderici2016}), image regions
near masked bits result in highly visible blocking artifacts
(Figure~\ref{fig:results-est-example-iter0}).  In addition to the
reconstruction-error term, we add a penalty for non-zero bits in our
codes: this helps bias our code distributions towards zero bits
(Figure~\ref{fig:bit-distributions}), making entropy compression more
effective.

The training approach that we found to work best for creating SCT
networks is to create a combined loss function from two interleaved
reconstruction passes: one pass to train for accurate reconstructions
and one pass to train for stop-code tolerance.  Without the SCT pass,
the networks would not see masked codes and code trimming until very
late in the training process.  By that point in training, they are
less able to adjust to the masked codes without the types of artifacts
shown in Figure~\ref{fig:results-est-example-iter0}-a: they already
have effectively made symbol assignments at the bottleneck/binarizer
that can not conform to an all-zero stop code.

To allow this two-pass training, for each mini-batch in the training,
just before the $k^{th}$ iteration, we record $S^{k-1}$ and $C^{k-1}$,
the `state' and `cell' vectors for all of the decoder's LSTM units and
$E_{\max}^{k-1}$ and $E_{\min}^{k-1}$, the maximum and minimum
reconstruction errors on that mini batch.
We then run the $k^{th}$,
iteration twice, by restoring the LSTM memory to $S^{k-1}$ and
$C^{k-1}$ between the two passes at the $k^{th}$ iteration.

\begin{figure}
  \centering
  \includegraphics[width=\linewidth]{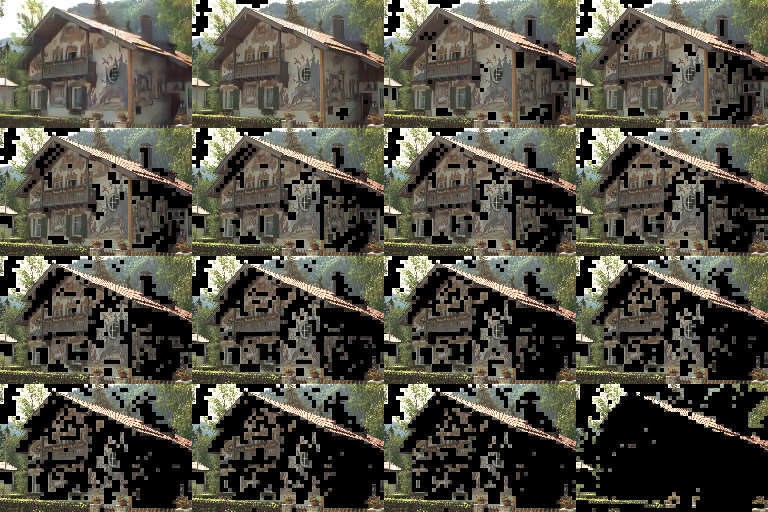}
  \caption{\small Visualization of coding progress in SCT method.
    Each image represents a new iteration.
    Trimmed codes are represented by masked portions of the
    image. Areas where the $L_1$ error remains above threshold
  continue to transmit codes, up to our iteration limit.}
  \label{fig:masks}
\end{figure}

First, we introduce a SCT-training pass through each iteration in
order to allow the neural network
to learn on examples with masked codes.  To present
the network with reasonable examples of masked codes, we set an
artificially high masking threshold, creating a ‘fake’ mask that will
always have some masked-off areas.  In the masked-off areas, we reset all of
the bits that will be seen by the decoder to zero, just as they would
have been if the mask were generated naturally, with the regular
masking threshold.  We set the artificially high masking threshold to
$(k/K * (E_{\max}^{k-1} - E_{\min}^{k-1}) +
E_{\min}^{k-1})$
where $K$ is the maximum number of iterations over which we are
training.
We add the $L_1$ reconstruction error of the output generated by these
forced-mask iterations to the reconstruction-error loss from the
``natural-mask'' reconstruction (the second pass).  Note that the
forced-mask 
reconstruction
builds on the natural-mask reconstruction for all the
iterations prior to the current iteration.
\begin{figure*}
  \centering
  \begin{minipage}{0.245\textwidth}
  \centering
  \includegraphics[width=\linewidth]{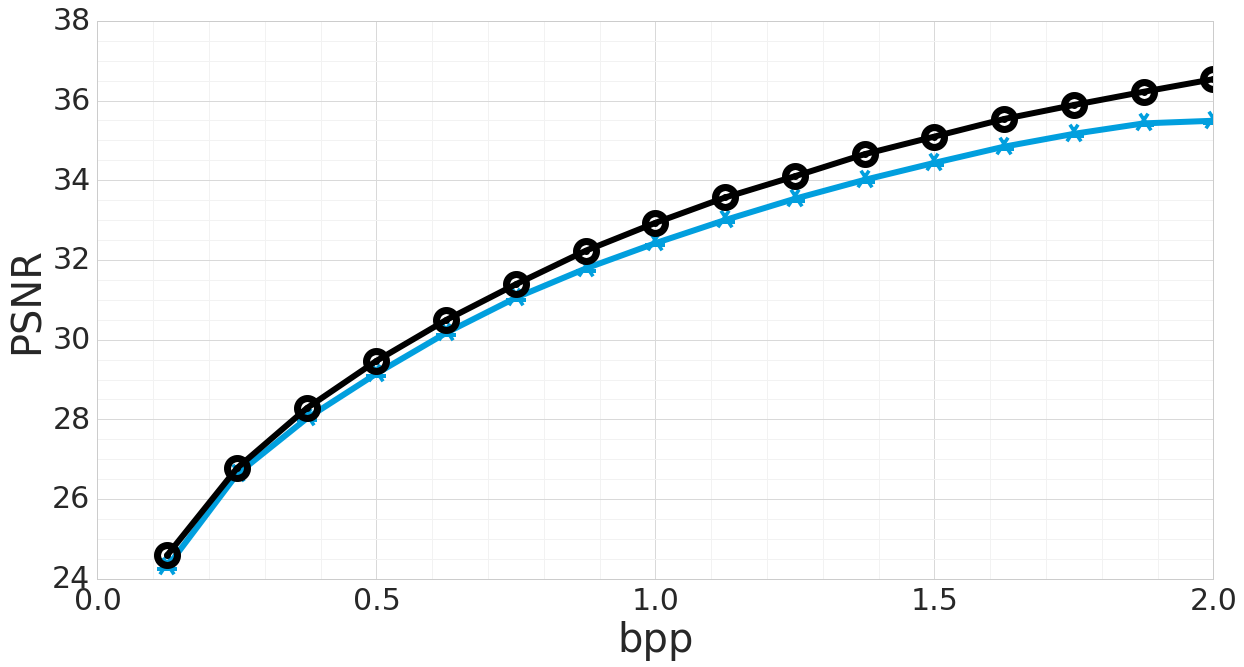}
  \linebreak \small
  (a) PSNR vs nominal bitrate
  \end{minipage}
  \hfill
  \begin{minipage}{0.245\textwidth}
  \centering
  \includegraphics[width=\linewidth]{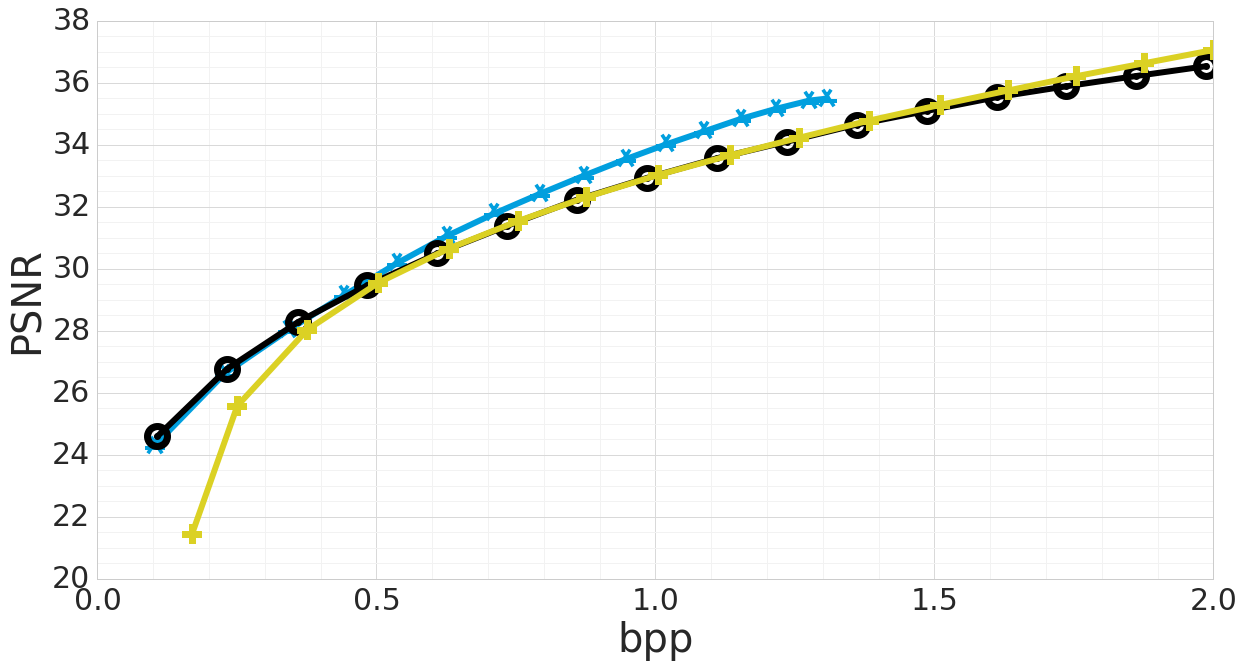}
  \linebreak \small
  (b) PSNR vs true bitrate
  \end{minipage}
  \hfill
  \begin{minipage}{0.245\textwidth}
  \centering
  \includegraphics[width=\linewidth]{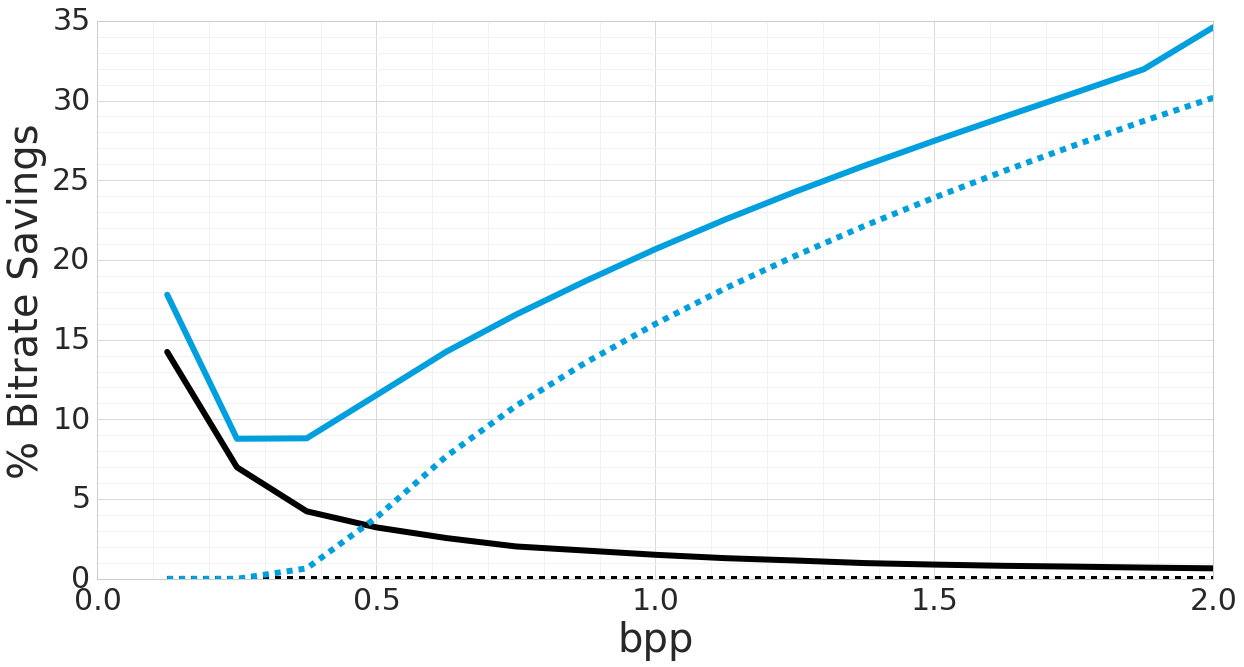}
  \linebreak \small
  (c) Bitrate savings from LZ77 and~code trimming
  \end{minipage}
  \hfill
  \begin{minipage}{0.245\textwidth}
  \centering
  \includegraphics[width=\linewidth]{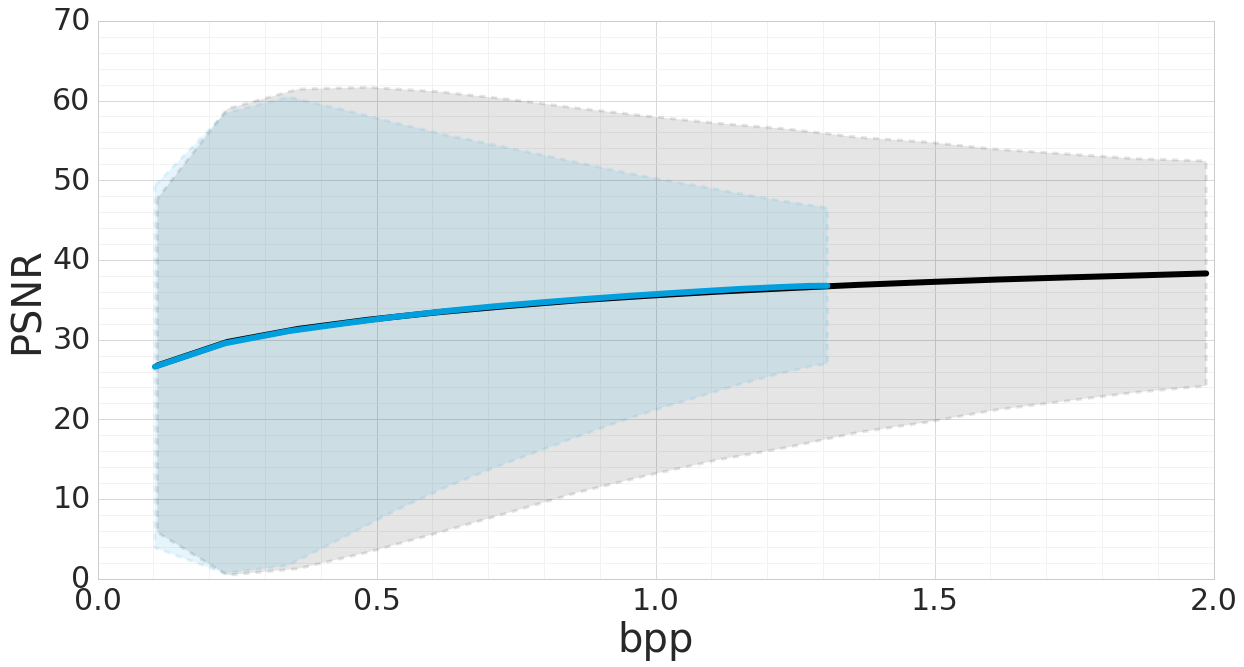}
  \linebreak \small
  (d) Within-image quality variance
  \end{minipage}
  \centering
 \caption{\small This figure shows results on the Kodak dataset, with the
   black lines for non-SCT coding~\cite{Toderici2016} and blue lines
   for our SCT method: (a) PSNR Rate-distortion (RD) curve,
   using the {\em nominal} bitrate: neither
   stop-code trimming nor LZ77 is done, so the RD curve for the
   non-SCT coding is higher. (b) PSNR RD curve using LZ77 compression
   and stop-code trimming: black line for~\cite{Toderici2016} after
   LZ77; blue line for EST coding with
   stop-code trimming and LZ77; yellow line for JPEG420. (c) Percent
  bitrate savings for SCT coding
    (blue) and~\cite{Toderici2016} (black) over the nominal bitrate: hashed
    lines show bitrate savings from stop-code trimming; solid lines
    show savings including LZ77. (d) Reduction in within-image
    variance in reconstruction quality using SCT coding: gray shading
    shows within-reconstruction-image variance with~\cite{Toderici2016};
    blue shading shows same with EST coding.}
  \label{fig:results}
\end{figure*}

We then run a second pass through the iteration, after
restoring $S^{k-1}$ and $C^{k-1}$, so that the
previous (forced-masking) pass through the $k^{th}$ iteration does not
impact our results.  In this second pass, we
use the masks generated ‘naturally’ by the expanding the previous
iteration’s (natural) mask according to the stop codes present in the
current iteration’s (natural) binary codes.  This reconstruction is
used to form the main reconstruction error (and its corresponding
loss), just as it was in~\cite{Toderici2016}.   It is the $S^k$,
$C^k$, $E_{\max}^k$ and $E_{\min}^k$ values from this ‘natural’
decoding that are recorded before the next iteration is run.

\section{RESULTS}
\label{sec:results}

Figures~\ref{fig:results-est-example-iter0} and~\ref{fig:results-75bpp}
provide examples of the difference between non-SCT and SCT
reconstructions, for 0.12 bpp and for 0.75 bpp true bit rate (that is,
the bitrate after LZ77/stop-trimming).
At 0.12 bpp, the non-SCT reconstruction is sharper than the SCT
reconstruction (see the vertical edges in Figure~\ref{fig:results-est-example-iter0}-b) but the quality
of the SCT reconstruction is more uniform than the non-SCT
reconstruction (Figure~\ref{fig:results-est-example-iter0}-a).  At 0.75 bpp, the SCT reconstruction is nearly
uniformly better than the non-SCT reconstruction (see
Figure~\ref{fig:results-75bpp}-c) and is a more uniform quality
across the reconstruction.

Figure~\ref{fig:masks} shows how the
SCT reconstruction progresses across iterations.  Almost a third of
the codes are trimmed before the $10^{th}$ iteration.  More than 80\% of the
codes are trimmed before the $16^{th}$ iteration on this image.

Figure~\ref{fig:results}-a and~\ref{fig:results}-b show the rate-distortion ({\em RD}) curves
for SCT and non-SCT~\cite{Toderici2016} coding, using the {\em
  nominal} bitrate and the bitrate with LZ77/stop-trimming,
respectively.  If the nominal bitrate is used, SCT coding
does worse than non-SCT coding.
This is expected since
the SCT architecture is required to balance
reconstruction quality
with robustness to
code trimming (to provide SCT) and reduction in bit-level
entropy (to improve LZ77).  Figure~\ref{fig:results}-b shows that the
SCT more than makes up for this nominal loss, once stop-code trimming
and LZ77 are used to be provide a more accurate RD curve.

In Figure~\ref{fig:results}-c, we show the amount of gain due to LZ77
on both non-SCT and SCT codes.  We also show the gains due to trimming
stop
codes (that is, all zeros) after the transmission of the first
instance of that in a given tile.  With the non-SCT codes, we see less
than 1\% additional bitrate compression using LZ77 at the higher bit
rates.  In contrast, we see a 35\% bitrate reduction in SCT codes over
the nominal bitrates: about 30\% due to stop-code trimming and the
last 5\% due to LZ77.

To compare the uniformity of the image quality within a
reconstruction, we took a reconstruction from the non-SCT architecture
at 0.745 bpp (after LZ77)
and a reconstruction from the SCT architecture that was closest to
that same bitrate (after stop-code
trimming and LZ77) and compared the $L_1$ error in each $32 \times 32$
section of the image.  Figure~\ref{fig:results-75bpp}-c shows
black where the SCT reconstruction
is worse than the non-SCT reconstruction and white were it is better.
The image areas where SCT reconstruction is worse than non-SCT are
those areas where the reconstructions are already quite good, so
stop coding has been used (compare Figure~\ref{fig:results-75bpp}-c to
the first and second line of Figure~\ref{fig:masks}).  The
$L_1$ tile-error mean and, more importantly, standard deviations are
both 11\% lower for the EST reconstruction than for the non-EST
reconstruction: for near 0.75 bpp on this image, it is 4.7 vs 5.3 for
the mean and 4.3 vs 4.8 for the
standard deviation.

\section{CONCLUSIONS}
\label{sec:conclusions}

This work has introduced a training method to allow neural-network
based compression systems to adaptively vary the number of symbols
transmitted for different parts of the compressed image, based on the local
underlying content.  Our method
allows the networks to remain fully convolutional, thereby
avoiding the blocking artifacts that often accompany block-based
approaches at low bit rates.  We have shown promising results with
this approach on a recursive auto-encoder structure which was
originally reported in~\cite{Toderici2016}.

There are many interesting directions to extend the current work.
The most immediate would be to train and apply neural-network-based
entropy coding~\cite{Toderici2016}, to exploit
the known spatial structure.  Longer term, one
interesting area is to consider changing the stop-code threshold on a
per-iteration basis, to encourage even more uniform quality
distributions. Another research direction is to take this same
SCT idea 
and apply it to non-recurrent auto-encoders.  This would require
assigning ``priority'' to different sections of the auto-encoder's
bottleneck, as well as determining the best stop-code structure.


%
%
%


\pagebreak

\balance

\bibliographystyle{IEEEbib}
\bibliography{biblio}

\begin{thebibliography}{10}

\bibitem{BalleLS16a}
J.~Ball{\'{e}}, V.~Laparra, and E.~Simoncelli,
\newblock ``End-to-end optimized image compression,''
\newblock {\em CoRR}, vol. abs/1611.01704, 2016.

\bibitem{Toderici2016}
G.~Toderici, D.~Vincent, N.~Johnston, {S.J.} Hwang, D.~Minnen, J.~Shor, and
  M.~Covell,
\newblock ``Full resolution image compression with recurrent neural networks,''
\newblock {\em CoRR}, vol. abs/1608.05148, 2016.

\bibitem{gregor2016conceptual}
K.~{Gregor}, F.~{Besse}, D.~{Jimenez Rezende}, I.~{Danihelka}, and
  D.~{Wierstra},
\newblock ``{Towards Conceptual Compression},''
\newblock {\em ArXiv e-prints}, 2016.

\bibitem{JPEG}
W.~Pennebaker and J.~Mitchell,
\newblock {\em {JPEG}: Still Image Compression Standard},
\newblock Kluwer Academic Publishers, 1992.

\bibitem{WebP}
Google,
\newblock ``{WebP}: Compression techniques
  ({http://developers.google.com/}\-{speed/}\-{webp/}\-{docs/}\-compression),''
  Accessed: 2017-01-30.

\bibitem{BPG}
F.~Bellard,
\newblock ``{BPG} image format ({http://bellard.org/}\-{bpg/}),'' Accessed:
  2017-01-30.

\bibitem{Hinton2006}
G.~E. Hinton and R.~R. Salakhutdinov,
\newblock ``Reducing the dimensionality of data with neural networks,''
\newblock {\em Science}, vol. 313, no. 5786, pp. 504--507, 2006.

\bibitem{Krizhevsky2011}
A.~Krizhevsky and G.~E. Hinton,
\newblock ``Using very deep autoencoders for content-based image retrieval,''
\newblock in {\em European Symposium on Artificial Neural Networks}, 2011.

\bibitem{vincent2010}
P.~Vincent, H.~Larochelle, Y.~Bengio, and P.-A. Manzagol,
\newblock ``Extracting and composing robust features with denoising
  autoencoders,''
\newblock {\em Journal of Machine Learning Research}, 2012.

\bibitem{Toderici2015}
G.~Toderici, S.~O'Malley, S.J. Hwang, D.~Vincent, D.~Minnen, S.~Baluja,
  M.~Covell, and R.~Sukthankar,
\newblock ``Variable rate image compression with recurrent neural networks,''
\newblock {\em ICLR 2016}, 2016.

\bibitem{Minnen2017}
D.~Minnen, G.~Toderici, M.~Covell, T.~Chinen, N.~Johnston, J.~Shor, S.J. Hwang,
  D.~Vincent, and S.~Singh,
\newblock ``Spatially adaptive image compression using a tiled deep network,''
  submission to ICIP 2017.

\bibitem{tensorflow}
M.~Abadi, A.~Agarwal, P.~Barham, E.~Brevdo, Z.~Chen, C.~Citro, G.~Corrado,
  A.~Davis, J.~Dean, M.~Devin, S.~Ghemawat, I.~Goodfellow, A.~Harp, G.~Irving,
  M.~Isard, Y.~Jia, R.~Jozefowicz, L.~Kaiser, M.~Kudlur, J.~Levenberg,
  D.~Man\'{e}, R.~Monga, S.~Moore, D.~Murray, C.~Olah, M.~Schuster, J.~Shlens,
  B.~Steiner, I.~Sutskever, K.~Talwar, P.~Tucker, V.~Vanhoucke, V.~Vasudevan,
  F.~Vi\'{e}gas, O.~Vinyals, P.~Warden, M.~Wattenberg, M.~Wicke, Y.~Yu, and
  X.~Zheng,
\newblock ``{TensorFlow}: Large-scale machine learning on heterogeneous systems
  (http://tensorflow.org),'' 2015,
\newblock Software available from tensorflow.org.

\bibitem{gzip}
J.-L. Gailly,
\newblock ``{GNU} {Gzip}: General file (de)compression
  ({http://www.gnu.org/}\-{software/}\-{gzip/}\-{manual/}\-{gzip.html}),''
  Accessed: 2017-01-30.

\end{thebibliography}

\end{document}